\documentclass{article}
 \usepackage{graphicx} 
\usepackage{amsmath}
\usepackage{amssymb} 
\usepackage{wrapfig}
\usepackage{color}
\usepackage{soul}
\usepackage{ulem}
\usepackage{multirow}
\usepackage{float}
\usepackage{subcaption}
\usepackage{svg}
\usepackage{booktabs}
\usepackage{hyperref}
\hypersetup{
    colorlinks = true,
    urlcolor = blue,
    citecolor    = blue,
    linkcolor    = blue
    }

\usepackage[sort,numbers]{natbib}

\newcommand\Tstrut{\rule{0pt}{2.6ex}}         
\newcommand\Bstrut{\rule[-1.3ex]{0pt}{0pt}}   

\usepackage{geometry}

\title{Geometric Multi-color Message Passing Graph Neural Networks for Blood-brain Barrier Permeability Prediction}
\author{
Trung Nguyen$^{1}$\thanks{These authors contributed equally to this work.}\,,\ 
Md Masud Rana$^{2}$\footnotemark[1]\,,\ 
Farjana Tasnim Mukta$^{2}$,\ 
Chang-Guo Zhan$^{3}$, \\
and\ 
Duc Duy Nguyen$^{4}$\footnote{Address correspondence to Duc Duy Nguyen. E-mail: ducnguyen@utk.edu} \\
$^1$The Bredesen Center, University of Tennessee, Knoxville, TN 37996, USA \\
$^2$Department of Mathematics, Kennesaw State University, Kennesaw, GA 30144, USA \\
$^3$Department of Pharmaceutical Sciences, University of Kentucky, Lexington, KY 40506, USA\\
$^4$Department of Mathematics, University of Tennessee, Knoxville, TN 37996, USA
}
\begin{document}

\maketitle

\begin{abstract}
    Accurate prediction of blood-brain barrier permeability (BBBP) is essential for central nervous system (CNS) drug development. While graph neural networks (GNNs) have advanced molecular property prediction, they often rely on molecular topology and neglect the three-dimensional geometric information crucial for modeling transport mechanisms. This paper introduces the geometric multi-color message-passing graph neural network (GMC-MPNN), a novel framework that enhances standard message-passing architectures by explicitly incorporating atomic-level geometric features and long-range interactions. Our model constructs weighted colored subgraphs based on atom types to capture the spatial relationships and chemical context that govern BBB permeability. We evaluated GMC-MPNN on three benchmark datasets for both classification and regression tasks, using rigorous scaffold-based splitting to ensure a robust assessment of generalization. The results demonstrate that GMC-MPNN consistently outperforms existing state-of-the-art models, achieving superior performance in both classifying compounds as permeable/non-permeable (AUC-ROC of 0.947 and 0.9212) and in regressing continuous permeability values (RMSE of 0.5628, Pearson correlation of 0.6947). An ablation study further quantified the impact of specific atom-pair interactions, revealing that the model's predictive power derives from its ability to learn from both common and rare, but chemically significant, functional motifs. By integrating spatial geometry into the graph representation, GMC-MPNN sets a new performance benchmark and offers a more accurate and generalizable tool for drug discovery pipelines.
\end{abstract}

\vspace{1em}

\section{Introduction}

The blood-brain barrier (BBB) is a highly selective interface that regulates the exchange of substances between the bloodstream and the brain, ensuring central nervous system (CNS) homeostasis by preventing the entry of harmful compounds while allowing essential nutrients and gases to pass through \cite{daneman2015blood, kadry2020blood}. BBB permeability (BBBP) is tightly controlled by the endothelial cells that form the barrier, whose tight adhesion and adhesion junctions restrict the passage of large or hydrophilic molecules, allowing only small, lipophilic, or actively transported compounds to cross \cite{kadry2020blood}. 

Predicting BBB permeability is critical for CNS drug development. Compounds with insufficient permeability of BBB can fail to achieve therapeutic concentrations in the brain \cite{vilella2015endocytosis, gao2013targeted, dong2018current}. BBBP data can be broadly categorized into numerical and categorical types. Numerical data quantify a compound’s permeability by comparing its concentration in the brain to its concentration in the blood, providing a continuous scale for measuring how effectively a substance crosses the BBB. In contrast, categorical data offer a binary classification, labeling compounds as either BBB$^+$ (permeable) or BBB$^-$ (non-permeable), which simplifies evaluation by indicating whether a substance can penetrate the barrier.



Various in vivo and in vitro studies have been conducted to assess BBB permeability in drug molecules. In vivo, models provide physiologically relevant information but are resource-intensive, time-consuming, and ethically challenging \cite{bagchi2019vitro}. In vitro models, including coculture and dynamic systems \cite{cucullo2011dynamic}, offer controlled environments for the testing of compounds but still involve significant costs and time. To address these limitations, in silico methods have gained prominence as efficient alternatives for BBBP prediction. These models can be broadly categorized into two main types: traditional quantitative structure-activity relationship (QSAR) models and machine learning-based models. 

QSAR models are among the earliest computational approaches used to predict BBB permeability. These models establish statistical correlations between molecular descriptors (e.g., lipophilicity, molecular weight, and hydrogen bonding capacity) and experimentally determined BBB permeability. Early QSAR models relied on multiple linear regression (MLR) and partial least squares (PLS) to identify molecular features associated with BBB permeability \cite{patel2014quantitative,faramarzi2022development,platts2001correlation}. Some QSAR models use physicochemical rules, such as Lipinski's Rule of Five or Veber's Rule\cite{lipinski2000drug,sakiyama2021prediction}, to estimate whether a compound is likely to penetrate the BBB. While these QSAR models provide interpretability, they often struggle with complex, nonlinear relationships in molecular data, leading to reduced predictive accuracy for diverse chemical compounds \cite{faramarzi2022development}.

With the availability of large molecular datasets \cite{b3db,wu2017moleculenet}, machine learning (ML) models have emerged as a more flexible alternative to QSAR-based methods. These models improve predictive performance by capturing non-linear relationships between molecular features and BBB permeability.  Support vector machine (SVM), Random forest (RF), and gradient boosting trees (GBT) algorithms are particularly effective in handling high-dimensional molecular data, integrating a vector of molecular descriptors to enhance prediction accuracy \cite{kumar2021b3pred, sakiyama2021prediction,kumar2022deepred,yuan2018improved}. Deep learning architectures have further advanced this field by automating feature extraction. For instance, convolutional neural networks (CNNs) are applied to 2D molecular images, learning spatial patterns associated with BBB permeability \cite{tang2022merged}. Whereas recurrent neural networks (RNNs) and long short-term memory (LSTM) networks process SMILES representations to model molecular properties in a sequential manner \cite{alsenan2020recurrent}. Additionally, hybrid approaches such as the DeePred-BBB Model \cite{kumar2022deepred} and LightBBB framework \cite{shaker2021lightbbb} incorporate various descriptor-based features, including physicochemical properties and molecular fingerprints, to enhance classification performance. Further advancements have also explored multi-task learning frameworks, which simultaneously predict several properties, and strategies involving the fusion of multi-level features to create more comprehensive molecular representations \cite{yang2025multitask, yang2025multitasktoxicity}.


In recent years, graph neural networks (GNNs) have emerged as a promising alternative by shifting from descriptor-based input representations to direct processing of molecular graphs \cite{wu2020comprehensive}. Unlike traditional ML/DL models that rely on precomputed molecular fingerprints or an array of descriptors, GNNs represent molecules as graphs, where atoms are treated as nodes and chemical bonds as edges, enabling the direct extraction of structural and relational features \cite{kipf2017semisupervised}. Message passing neural networks (MPNNs) \cite{gilmer2017neural} have been widely applied in this context, as these architectures allow information to be exchanged between atomic neighbors through iterative message-passing steps. GSL-MPP \cite{gslmpp} extends standard MPNNs by integrating graph structure learning on a molecular similarity graph, utilizing both intra- and inter-molecular relationships. CoMPT \cite{compt} introduces a node-edge message interaction mechanism within a transformer-based framework to explicitly capture edge information. It also employs a message diffusion strategy to prevent over-smoothing and enhance long-range dependencies. CD-MVGNN \cite{cdmvgnn} introduces a multi-view GNN that equally considers atoms and bonds by using dual encoders and a cross-dependent message-passing scheme for bidirectional information exchange.



While GNN-based models have significantly advanced BBBP prediction, their principal reliance on molecular topology constrains their ability to fully capture the structural and physicochemical properties governing blood-brain barrier permeability. These models often overlook the role of three-dimensional geometric information, which is critical for accurately modeling molecular interactions and transport mechanisms. Although some 3D-aware GNN architectures have been introduced for molecular property prediction, these models do not integrate long-range interactions between different atom types. Moreover, such architectures are highly complex, requiring multiple network layers and intricate integration mechanisms.

In this study, we introduce a novel geometric multi-color message-passing graph neural network (GMC-MPGNN) that extends the capabilities of standard MPNNs by incorporating atomic-level geometric features. Unlike conventional GNNs that operate purely on molecular connectivity, our proposed model integrates spatial representations and captures multi-range molecular interactions through the construction of multi-color subgraphs. Beyond classification tasks in BBBP, we extend the application of GMC-MPGNN to regression tasks using the B3DB dataset \cite{b3db}. Our results demonstrate that GMC-MPGNN outperforms existing state-of-the-art deep learning models, setting a new benchmark for BBBP modeling.

\section{Methodology}
\subsection{ Weighted Colored Subgraph Representation}
In this work, we first represent molecules through weighted colored subgraphs (WCS) to capture the geometric and chemical properties critical for accurately predicting molecular interactions and permeability. To this end, we model the molecule of interest via a molecular graph $\mathcal{G(V,E)}$, where vertices $\mathcal{V}$ denote the atoms, colored by their atom types. Edges $\mathcal{E}$ represent the non-covalent interactions between atoms. For a given molecular dataset, we collect a set $\mathcal{C}$ containing all the distinct atom types. We denote $\mathcal{C}_k$ as a $k^{\text{th}}$ atom type in $\mathcal{C}$. Formally, vertices $\mathcal{V}$ can be expressed as:

\begin{align}
    \mathcal{V} = \{ (\mathbf{r}_i, \alpha_i) | \mathbf{r}_i \in \mathbb{R}^3, \alpha_i \in \mathcal{C};\, i = 1, 2, \cdots, N \},
\end{align}

where $\mathbf{r}_i$ and $\alpha_i$ are the 3D coordinate and atom type for the $i^{\text{th}}$ atom, respectively. We define edge $\mathcal{E}$ using a weighted function $\Phi$, typically selected as a generalized exponential or Lorentz function \cite{nguyen2017rigidity}, which captures the decay of interaction strength as interatomic distances increase:
\begin{align}
\mathcal{E}=\left\{\Phi\left(\left\|\mathbf{r}_i-\mathbf{r}_j\right\| ; \eta_{kk'}\right) \mid \alpha_k=\mathcal{C}_k, \alpha_j = \mathcal{C}_{k'} ; i, j=1,2, \ldots, N\right\},
\end{align}

where $\left\|\mathbf{r}_i-\mathbf{r}_j\right\|$ denotes the Euclidean distance between atoms $i$ and $j$, and $\eta_{kk'}$ is the characteristic distance, defined as  {$\eta_{kk'}=\tau\left(r_{\text{vdw}_k}+r_{\text{vdw}_{k
'}}\right)$, with $r_\text{vdw}$ representing the van der Waals radius and $\tau$ is the scaling factor. This parameter allows the model to tune its focus on different interaction ranges. A value of $\tau < 1$ prioritizes interactions at distances shorter than the sum of vdW radii, crucial for capturing strong, specific non-covalent forces like hydrogen bonds. Conversely, $\tau > 1$ enables the model to learn from weaker, long-range electrostatic and van der Waals interactions. Common choices for $\Phi$ include:

\begin{align}
\Phi_E\left(\left\|\mathbf{r}_i-\mathbf{r}_j\right\| ; \eta_{\alpha_i \alpha_j}\right)=e^{-\left(\left\|\mathbf{r}_i-\mathbf{r}_j\right\| / \eta_{\alpha_i \alpha_j}\right)^\kappa}, \quad \kappa>0,
\end{align}

where $\kappa$ is the power parameter of the kernel, which helps approximate the ideal low-pass filter (ILF) \cite{nguyen2017rigidity}. A high $\kappa$ value creates a sharp, steep decay, effectively filtering interactions to a very specific distance range. A low $\kappa$ value creates a ``soft,'' gradual decay, allowing a wider range of interactions to contribute. By optimizing $\kappa$, the model learns the most effective ``shape'' of the interaction potential.

We now construct a multi-color subgraph $\mathcal{G}_{kk'}(\mathcal{V}_{kk'},\mathcal{E}_{kk'})\subset \mathcal{G(V,E)}$, for a pairwise atom types $\mathcal{C}_k$ and $\mathcal{C}_{k'}$ with $\mathcal{C}_k, \mathcal{C}_{k'} \in \mathcal{C}$. Conceptually, this WCS approach creates separate subgraphs for specific types of interactions—for example, mapping how all nitrogen atoms interact with all oxygen atoms across the entire molecule. The strength (weight) of these interactions in $\mathcal{E}_{kk'}$ depends on their spatial proximity and will allow the model to learn chemical properties that a simple connectivity graph would miss.

The vertices $\mathcal{V}_{kk'} \subset \mathcal{V}$ correspond to all atoms of types $\mathcal{C}_k$ and $\mathcal{C}_{k'}$ in the original molecule. We consider two cases for the edge set $\mathcal{E}_{kk'}$:
\begin{itemize}
    \item Subgraph of two distinct atom types ($k\neq k'$): The edge set $\mathcal{E}_{kk'}$ forms the interaction only between these two atom types, resulting in a bipartite graph.
    
    \item Subgraph of two same atom types ($k=k'$): This scenario leads to a fully connected graph where every two vertices form an interaction.
\end{itemize}

From these subgraphs, we compute intermediate matrices representing their structure. The detailed mathematical construction of the weighted adjacency ($A_{kk'}$) and Laplacian ($L_{kk'}$) matrices, along with the specific cutoff criteria for excluding covalent bonds, are provided in the Supporting Information (Section Methodological Details). Figure~\ref{fig:subgraph} illustrates this entire subgraph generation process.

\begin{figure}[!htb]
    \centering
    \includegraphics[width=1\textwidth]{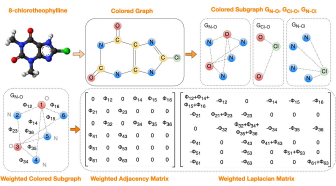}
    \caption{An illustration of the construction of weighted adjacency and weighted Laplacian matrices from the weighted colored subgraphs of a molecule. In the top row, an example molecule, 8-chlorotheophylline (\(\text{C}_7\text{H}_7\text{Cl}\text{N}_4\text{O}_2\); CHEBI:59771), the colored graph structure of the 8-chlorotheophylline, and three example colored subgraphs of \( G_{N-O}, G_{Cl-O}, G_{N-Cl} \). In the bottom row, generated weighted adjacency matrix (A) and weighted Laplacian matrix (L) from a example subgraph \( G_{N-O} \).}
    \label{fig:subgraph}
\end{figure}

Figure \ref{fig:subgraph} demonstrates the generation of the multi-color subgraph $\mathcal{G}_{\text{NO}}$ for 8-chlorotheophylline  (\(\text{C}_7\text{H}_7\text{Cl}\text{N}_4\text{O}_2\)). Figure \ref{fig:subgraph} also illustrates the construction of its weighted adjacency and weighted Laplacian matrices.

\subsection{Weighted Colored Subgraph-based Atomic Features}
We propose constructing atomic features $\mathcal{X}_i^{\mathrm{WCS}}$ for an atom $\left(\mathbf{r}_i, \alpha_i\right)$, where $\alpha_i=\mathcal{C}_k$, using its twelve corresponding weighted colored subgraphs $\mathcal{G}_{k k'}$. 
Each component $\left(\mathcal{X}_i^{\mathrm{WCS}}\right)_{k'}$ encodes the geometric influence of atom type $k'$ on the atom $\left(\mathbf{r}_i, \alpha_i\right)$ through the associated weighted colored subgraph $\mathcal{G}_{k k^{\prime}}$. To this end, we calculate the statistical measures, including summation, mean, median, minimum, maximum, and standard deviation, of all the geometric interactions in the subgraph $\mathcal{G}_{k k'}$ to atom $i$. Specifically
\begin{equation}
    \left(\mathcal{X}_i^{\mathrm{WCS}}\right)_{k'}=\left[(D_{k k'})_{i i}, \min_{j \in \mathcal{V}_{k k^{\prime}}}\left(A_{k k'}\right)_{i j}, \max_{j \in \mathcal{V}_{k k^{\prime}}}\left(A_{k k'}\right)_{i j}, \underset{j \in \mathcal{V}_{k k^{\prime}}}{\operatorname{mean}}\left(A_{k k'}\right)_{i j}, \underset{j \in \mathcal{V}_{k k^{\prime}}}{\operatorname{std}}\left(A_{k k'}\right)_{i j}\right],
\end{equation}

These terms quantify the cumulative interaction strengths from atoms of type $k'$ to atom $i$ and serve as a compact descriptor of atom-type-specific geometric context.

To complement these subgraph-based descriptors, we integrate cheminformatics atomic features (CAF), denoted as $\mathcal{X}_i^{\text{CAF}}$, derived from cheminformatics tools such as RDKit \cite{landrum2006rdkit}. These include the atom type, represented by a one-hot encoding of the atomic number (100 dimensions), the number of bonds the atom is involved in (6-dimensional one-hot encoding), formal charge (5-dimensional encoding), and chirality (4-dimensional encoding capturing cases such as unspecified, tetrahedral CW/CCW, or other). Additional descriptors include the number of bonded hydrogen atoms (5-dimensional), hybridization state (sp, sp2, sp3, sp3d, or sp3d2, encoded over 5 dimensions), aromaticity (a binary indicator of whether the atom is part of an aromatic system), and atomic mass (a real-valued scalar scaled by 1/100). All categorical features are one-hot encoded to ensure consistency and numerical stability in learning \cite{heid2023chemprop}.

In addition to atomic-level descriptors, we incorporate bond features, $\mathcal{X}_{ij}^{\text{Bond}}$, between atoms $i$ and $j$ to enrich the message-passing process with edge-level information \cite{heid2023chemprop}. These features, also derived from RDKit, include the bond type (single, double, triple, or aromatic, encoded in 4 dimensions), whether the bond is conjugated (1-dimensional), whether the bond is part of a ring (1-dimensional), and the bond stereochemistry (none, any, E/Z or cis/trans, encoded in 6 dimensions). All bond features are represented using one-hot encoding.

\subsection{Message-passing Graph Neural Network}
The proposed GMC-MPNN integrates atomic features derived from both geometric subgraph analysis and cheminformatics-based representations. These features are propagated through the molecular graph using a message-passing architecture to predict BBBP.

At the initialization step, each atom $i$ is assigned a comprehensive feature vector $h_i^0$, constructed by concatenating the WCS-based features $\mathcal{X}_i^{\text{WCS}}$  with the cheminformatics atomic features $\mathcal{X}_i^{\text{CAF}}$
\begin{equation}
h_i^0 = \left[ \mathcal{X}_i^{\text{CAF}} \, \| \, \mathcal{X}_i^{\text{WCS}} \right].
\end{equation}
This richer initialization is essential for capturing more detailed structural information, which leads to improved property predictions. For edge-level encoding, each bond between atoms $i$ and $j$ is represented using the bond feature vector $\mathcal{X}_{ij}^{\text{Bond}}$, which includes bond type, conjugation, ring membership, and stereochemistry. These are used during message computation to condition the information exchanged between neighboring nodes.

During each message-passing iteration $t$, the representation of each atom is updated by aggregating information from its neighbors:
\begin{align}
m_i^{(t+1)} &= \sum_{j \in \mathcal{N}(i)} \phi \left( h_i^{(t)}, h_j^{(t)}, \mathcal{X}_{ij}^{\text{Bond}} \right), \\
h_i^{(t+1)} &= \tau \left( W_m h_i^{(t)} + W_u m_i^{(t+1)} \right).
\end{align}
Here, $\phi(\cdot)$ is a learnable message function that accounts for the sender's and receiver's current states and their connecting bond features. For example, we can choose $\phi(\cdot)$ as a multi-layer perceptron (MLP) applied to the concatenation of the input features. This allows the model to learn complex, non-linear interactions between atomic and bond features during message passing. $W_m$ and $W_u$ are learnable transformation matrices, and $\tau$ is a non-linear activation function (e.g., ReLU, PReLU, LeakyReLU, etc).

After $T$ message-passing iterations, we compute the final representation for each atom $i$ by aggregating the hidden states of its neighbors and combining them with the atom’s original feature vector:
\begin{equation}
h_i^{\text{final}} = \tau \left( W_a \left[ h_i^0 \, \| \, \sum_{j \in \mathcal{N}(i)} h_j^{(T)} \right] \right),
\end{equation}

where $W_a$ is a learnable weight matrix. $\sum_{j \in \mathcal{N}(i)} h_j^{(T)}$ represents the sum of the messages passed from all neighboring atoms \(j\) to atom \(i\) after \(T\) iterations, $h^0_i$ is the original atom-level feature for atom \(i\) to allow the atom's initial features to remain part of the final representation.

The final molecular representation $h_G$ is obtained by summing the final node representations:
\begin{equation}
h_\mathcal{G} = \sum_{i \in \mathcal{V}} h_i^{\text{final}}.
\end{equation}

This graph-level embedding captures both the local atomic environments and long-range geometric interactions derived from weighted colored subgraphs. It is passed through a feed-forward neural network (FFN)
\begin{equation}
\hat{y} = f(h_G).
\end{equation}

where $f$ is an FNN that outputs the final BBBP property, either as a probability for classification or a continuous value for regression. An illustration of the GMC-MPNN framework is shown in Figure \ref{fig:ggl_chem}.

This message-passing framework enables the model to integrate spatial subgraph descriptors, bond-level interactions, and standard chemical features into a unified graph-based learning paradigm. The GMC-MPNN architecture thus provides a scalable and interpretable approach for molecular property prediction, with demonstrated effectiveness on both BBBP classification and regression benchmarks.

\begin{figure}[!htb]
    \centering
    \includegraphics[width=1.1\textwidth]{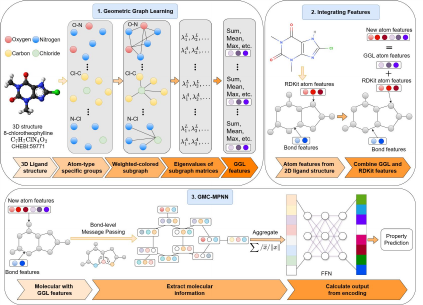}
    \caption{An illustration of GMC-MPNN atom-level fusion graph model. 1) Construct geometric graph learning atom features by considering statistical information (sum, mean, median, etc.) about the rigidity of the molecular graphs. 2) Converts molecular SMILES string to molecular graph using RDKit, and integrates with GGL features as new atom-level features. 3) Pass these combined features through a message-passing neural network to update all feature vectors, followed by an aggregation function and a feed-forward neural network for property prediction.}
    \label{fig:ggl_chem}
\end{figure}

\section{Datasets Preparations}

In this work, we utilize three distinct datasets related to BBBP. For clarity throughout the manuscript, we refer to them as $BBBP_{\text{cls}}^{\text{MolNet}}$ and $BBBP_{\text{cls}}^{\text{B3DB}}$ for the two classification datasets, and $BBBP_{\text{reg}}^{\text{B3DB}}$ for the regression dataset. For all datasets, we first performed a cleaning step on the SMILES strings to remove isolated ions and ensure chemical validity.

The $BBBP_{\text{cls}}^{\text{MolNet}}$ dataset contains 2,039 compounds with pre-existing 3D structures. The larger $BBBP_{\text{cls}}^{\text{B3DB}}$ and $BBBP_{\text{reg}}^{\text{B3DB}}$ datasets, derived from the B3DB benchmark, did not include 3D coordinates. For these, we generated a single, low-energy 3D conformer for each molecule using a hierarchical pipeline. This process was primarily conducted with the OpenEye OMEGA Toolkit, with a fallback procedure using RDKit and OpenBabel to maximize coverage. After removing molecules that could not be processed, we obtained final datasets of 7,740 molecules for classification and 1,047 for regression. A full description of the dataset preparation is available in the Supplemental Information~\cite{SI_GMC_MPNN}

A summary of the datasets utilized in this study is provided in Table~\ref{table:bbbp_summary_datasets}.

\begin{table}[H]
\begin{center}
\caption{Summary of datasets used in this study}
\label{table:bbbp_summary_datasets}
\begin{tabular}{l c c c c c}
	\hline
	    Dataset & Total Compounds & \# of BBB+ & \# of BBB– & Dataset Type \Tstrut\Bstrut\\
	\hline 
        $BBBP_{\text{cls}}^{\text{MolNet}}$ & 2,039 & 1,560 & 479 & Classification \Tstrut\\ 
        $BBBP_{\text{cls}}^{\text{B3DB}}$ & 7,740 & 4,905 & 2,835 & Classification \Tstrut\\ 
        $BBBP_{\text{reg}}^{\text{B3DB}}$ & 1,047 & N/A & N/A & Regression \Bstrut\\ 
        \hline
\end{tabular}
\end{center}
\end{table}

To ensure a robust evaluation of our models, we employed a scaffold splitting method to partition each dataset into training, validation, and test sets. Scaffold splitting groups molecules by their core structures, helping to ensure structural diversity in the training set and providing a stringent test of generalizability on novel chemical scaffolds. We adopted an 8:1:1 ratio for all three datasets, as recommended in recent studies \cite{shen2023molecular, chan2019advancing}. This approach mimics real-world drug discovery scenarios, where models must predict the properties of previously unseen compounds.

For each dataset, we performed 5 scaffold-based splits using seed values from 0 to 4. Final model performance is reported as the average across all splits to ensure stability and robustness in our results.

\subsection{Evaluation Metrics}
In assessing the predictive performance of models trained on the BBBP datasets, we employ a set of evaluation metrics tailored to the specific characteristics of each dataset. Below is a comprehensive explanation of these metrics.

For the classification task with the $BBBP_{\text{cls}}^{\text{MolNet}}$ and $BBBP_{\text{cls}}^{\text{B3DB}}$ dataset, the primary evaluation metric is the Area Under the ROC Curve (AUC-ROC). AUC summarizes the model's ability to distinguish between positive and negative instances. To calculate AUC, we compute the True Positive Rate (TPR) and False Positive Rate (FPR), where TPR measures the proportion of actual positives correctly classified, and FPR measures the proportion of actual negatives incorrectly classified as positives. The ROC curve plots TPR against FPR, and AUC provides an aggregate measure of performance, with higher values indicating better classification ability.

For the $BBBP_{reg}^{\text{B3DB}}$ dataset, which is a regression task, we use the Root Mean Square Error (RMSE) and Pearson Correlation as the primary evaluation metrics. RMSE measures the average magnitude of the prediction errors, with lower values indicating better model performance. It is calculated as 
\begin{align}
    \text{RMSE} & = \sqrt{\frac{1}{n} \sum_{i=1}^{n} (y_i - \hat{y}_i)^2},
\end{align}
where \( y_i \) is the true value and \( \hat{y}_i \) is the predicted value.

Pearson Correlation measures the linear relationship between the predicted and true values. The formula for Pearson's correlation coefficient (\( r \)) is defined as follows

\begin{equation}
    r = \frac{\sum_{i=1}^{n} (x_i - \bar{x})(y_i - \bar{y})}{\sqrt{\sum_{i=1}^{n} (x_i - \bar{x})^2 \sum_{i=1}^{n} (y_i - \bar{y})^2}},
\end{equation}

where \( x_i \) and \( y_i \) are the values of the two variables (predicted and true values), \( \bar{x} \) and \( \bar{y} \) are the means of the predicted and true values, respectively, and \( n \) is the number of data points.

\section{Results and Discussion}
In this section, we present the experimental results of GMC-MPNN on the datasets $BBBP_{\text{cls}}^{\text{MolNet}}$, $BBBP_{\text{cls}}^{\text{B3DB}}$, and $BBBP_{reg}^{\text{B3DB}}$ coupled with various state-of-the-art models to evaluate its performance in BBBP.

\subsection{Classification Task Experiments}

We evaluate the classification performance and generalizability of our proposed GMC-MPNN model on two benchmark datasets: the MoleculeNet BBBP dataset ($BBBP_{\text{cls}}^{\text{MolNet}}$) and an extended version derived from the B3DB benchmark ($BBBP_{\text{cls}}^{\text{B3DB}}$). Both datasets are designed to predict blood-brain barrier permeability labels (BBB+ or BBB–) and are split using scaffold-based partitioning to simulate the realistic scenario of predicting molecular properties for structurally novel compounds.

\begin{figure}[h!]
    \centering
    \includegraphics[width=0.92\textwidth]{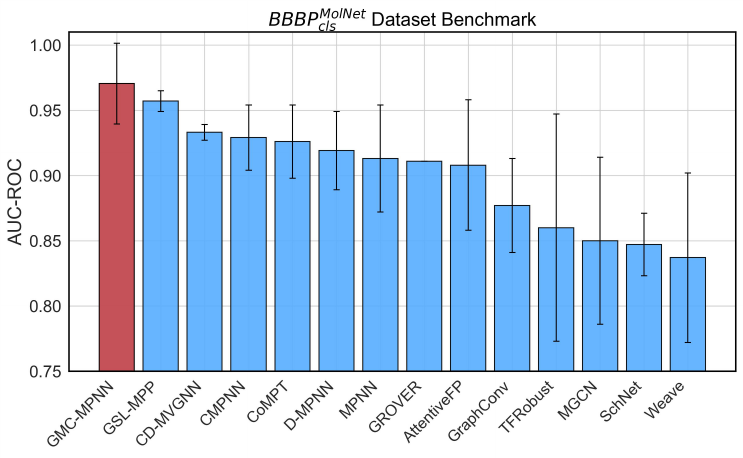}
    \caption{Performance comparison of different models on the $BBBP_{\text{cls}}^{\text{MolNet}}$ dataset. The red bars highlight our {GMC-MPNN} model.}
    \label{fig:BBBP_d1_top_models}
\end{figure}

\begin{table}[h!]
    \centering
    \caption{Performance of various models on $BBBP_{\text{cls}}^{\text{MolNet}}$ dataset}
    \begin{tabular}{@{}ll@{}}
        \toprule
        \textbf{Model} & \textbf{AUC-ROC} \\ \midrule
        \textbf{GMC-MPNN} & \textbf{0.947 $\pm$ 0.011} \\
        GSL-MPP & 0.935 $\pm$ 0.039 \\
        CD-MVGNN & 0.933 $\pm$ 0.006 \\
        CMPNN & 0.929 $\pm$ 0.025 \\
        CoMPT & 0.926 $\pm$ 0.028 \\
        D-MPNN & 0.919 $\pm$ 0.030 \\
        MPNN & 0.913 $\pm$ 0.041 \\
        GROVER & 0.911 \\
        AttentiveFP & 0.908 $\pm$ 0.050 \\
        GraphConv & 0.877 $\pm$ 0.036 \\
        TFRobust & 0.860 $\pm$ 0.087 \\
        MGCN & 0.850 $\pm$ 0.064 \\
        SchNet & 0.847 $\pm$ 0.024 \\
        Weave & 0.837 $\pm$ 0.065 \\
        \bottomrule
    \end{tabular}
    \label{tab:final_bbbp_benchmark}
\end{table}

On the $BBBP_{\text{cls}}^{\text{MolNet}}$ dataset, as illustrated in Figure~\ref{fig:BBBP_d1_top_models} and detailed in Table~\ref{tab:final_bbbp_benchmark}, GMC-MPNN achieves the best AUC-ROC score of $0.947 \pm 0.011$. This strong performance demonstrates the value of incorporating spatially-informed atomic features and long-range geometric interactions via weighted colored subgraphs. GSL-MPP~\cite{gslmpp} ($0.935 \pm 0.039$) and CD-MVGNN~\cite{cdmvgnn} ($0.933 \pm 0.006$) follow as the next-best models, while traditional message-passing models like MPNN~\cite{gilmer2017neural} ($0.913 \pm 0.041$) and GraphConv~\cite{graphconv} ($0.877 \pm 0.036$) perform less competitively.} 

\begin{figure}[h!]
    \centering
    \includegraphics[width=0.92\textwidth]{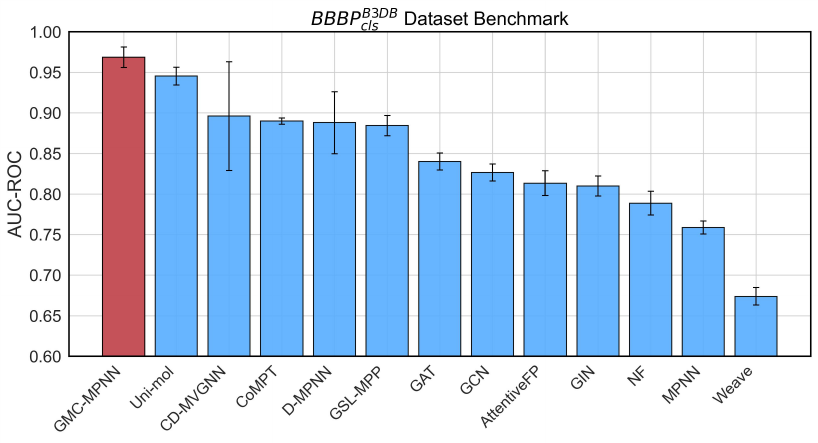}
    \caption{Performance comparison of different models on $BBBP_{\text{cls}}^{\text{B3DB}}$ dataset, evaluated using scaffold-balanced splits.}
    \label{fig:BBBP_cls_b3db}
\end{figure}

We next evaluate all models on the $BBBP_{\text{cls}}^{\text{B3DB}}$ dataset, which is considered more challenging due to its greater structural diversity and broader distribution of molecular scaffolds. As shown in Figure~\ref{fig:BBBP_cls_b3db} and detailed in Table~\ref{tab:final_bbbp_b3db_benchmark}, GMC-MPNN again achieves the best performance with an AUC-ROC of $0.9212 \pm 0.0261$, confirming its robustness in generalizing across these more varied structures. Following the top performers, a cluster of recent models demonstrates comparable results, including CD-MVGNN~\cite{cdmvgnn} ($0.8958 \pm 0.0669$), CoMPT~\cite{compt} ($0.8896 \pm 0.0039$), D-MPNN~\cite{heid2023chemprop} ($0.8878 \pm 0.0381$), and GSL-MPP~\cite{gslmpp} ($0.8840 \pm 0.0126$). These are followed by GAT~\cite{velickovic2017graph} and GCN~\cite{kipf2016semi}. Models like MPNN~\cite{gilmer2017neural} and Weave perform less favorably, with AUCs under 0.77, highlighting their limitations in capturing complex structural patterns under challenging generalization scenarios.

\begin{table}[h!]
    \centering
    \caption{Performance of various models on the $BBBP_{\text{cls}}^{\text{B3DB}}$ dataset, sorted by performance.}
    \begin{tabular}{@{}ll@{}}
        \toprule
        \textbf{Model} & \textbf{AUC-ROC} \\ \midrule
        \textbf{GMC-MPNN} & \textbf{0.9212 $\pm$ 0.0261} \\
        CD-MVGNN & 0.8958 $\pm$ 0.0669 \\
        CoMPT & 0.8896 $\pm$ 0.0039 \\
        D-MPNN & 0.8878 $\pm$ 0.0381 \\
        GSL-MPP & 0.8840 $\pm$ 0.0126 \\
        GAT & 0.8401 $\pm$ 0.0105 \\
        GCN & 0.8264 $\pm$ 0.0105 \\
        AttentiveFP & 0.8133 $\pm$ 0.0153 \\
        GIN & 0.8097 $\pm$ 0.0123 \\
        NF & 0.7886 $\pm$ 0.0146 \\
        MPNN & 0.7586 $\pm$ 0.0080 \\
        Weave & 0.6737 $\pm$ 0.0106 \\
        \bottomrule
    \end{tabular}
    \label{tab:final_bbbp_b3db_benchmark}
\end{table}

Overall, our GMC-MPNN model consistently outperforms all baselines on both benchmarks, achieving AUC-ROC scores of $0.947 \pm 0.011$ on the MoleculeNet version and $0.9212 \pm 0.0261$ on the B3DB variant. The superior and consistent performance of GMC-MPNN across both benchmarks confirms the utility of incorporating geometry-aware atomic and subgraph features. These results underscore the importance of evaluating molecular property prediction models on diverse and carefully curated datasets to ensure reliable generalization in drug discovery pipelines.

\subsection{Regression Task Experiments}

\begin{figure}[h!]
    \centering
    \includegraphics[width=\linewidth]{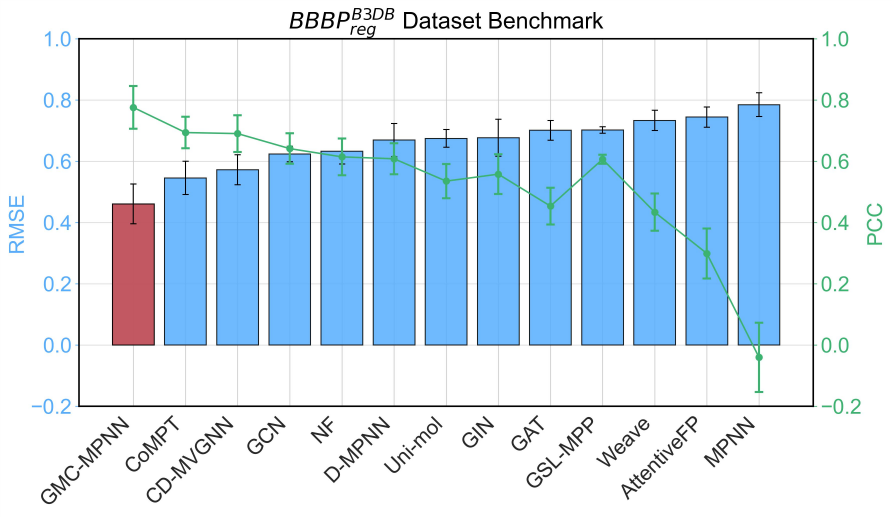}
    \caption{Performance comparison of different models on the $\text{BBBP}_{\text{reg}}^{\text{B3DB}}$ dataset with RMSE and Pearson Correlation scores. The red bars showcase the performances of our geometric graph learning-based model, GMC-MPNN. We compare against several baseline and state-of-the-art models: GCN~\cite{kipf2016semi}, NF~\cite{nf}, D-MPNN~\cite{heid2023chemprop}, GIN~\cite{genova2017graph}, GAT~\cite{velickovic2017graph}, Weave~\cite{kearnes2016molecular}, AttentiveFP~\cite{xiong2019pushing}, CoMPT~\cite{compt}, CD-MVGNN~\cite{cdmvgnn}, GSL-MPP~\cite{gslmpp}, and MPNN~\cite{gilmer2017neural}.}
    \label{fig:reg_benchmark}
\end{figure}

\begin{table}[h!]
    \centering
    \caption{Performance of various models on the $BBBP_{reg}^{\text{B3DB}}$ dataset, sorted by ascending RMSE. PCC stands for Pearson's correlation coefficient.}
    \begin{tabular}{@{}lcc@{}}
        \toprule
        \textbf{Model} & \textbf{RMSE} & \textbf{PCC} \\ \midrule
        \textbf{GMC-MPNN} & \textbf{0.5628$\pm$0.0651} & \textbf{0.6947$\pm$0.0515} \\
        CoMPT & 0.5749$\pm$0.0538 & 0.6828$\pm$0.0659 \\
        CD-MVGNN & 0.6246$\pm$0.0413 & 0.6731$\pm$0.0586 \\
        GCN & 0.6249$\pm$0.0252 & 0.6415$\pm$0.0498 \\
        NF & 0.6325$\pm$0.0412 & 0.6145$\pm$0.0598 \\
        D-MPNN & 0.6696$\pm$0.0536 & 0.6083$\pm$0.0504 \\
        GIN & 0.6773$\pm$0.0604 & 0.5581$\pm$0.0651 \\
        GAT & 0.7011$\pm$0.0321 & 0.4538$\pm$0.0599 \\
        GSL-MPP & 0.7021$\pm$0.0107 & 0.6067$\pm$0.0147 \\
        Weave & 0.7334$\pm$0.0330 & 0.4339$\pm$0.0609 \\
        AttentiveFP & 0.7444$\pm$0.0329 & 0.2992$\pm$0.0816 \\
        MPNN & 0.7850$\pm$0.0387 & -0.0400$\pm$0.1133 \\
        \bottomrule
    \end{tabular}
    \label{tab:bbbp_ggl_regression_benchmark}
\end{table}

We evaluate the performance of our proposed GMC-MPNN model on the regression task using the $BBBP_{\text{reg}}^{\text{B3DB}}$ dataset. GMC-MPNN, which incorporates geometry-aware WCS descriptors and max-pooling, achieves the best results among all evaluated models. It records the lowest RMSE of $0.5628 \pm 0.0651$ and the highest Pearson's correlation coefficient (PCC) of $0.6947 \pm 0.0515$. The strong performance demonstrates the effectiveness of WCS features in capturing both spatial and chemical contexts critical for BBB permeability prediction.

Table~\ref{tab:bbbp_ggl_regression_benchmark} and Figure~\ref{fig:reg_benchmark} present a comparative analysis across baseline models. Following GMC-MPNN, CoMPT~\cite{compt} emerges as the next best performer, achieving strong RMSE scores of $0.5749 \pm 0.0538$. Moderate results are observed from models such as CD-MVGNN~\cite{cdmvgnn}, GCN~\cite{kipf2016semi}, NF~\cite{nf}, D-MPNN~\cite{heid2023chemprop}, and GIN~\cite{genova2017graph}, record competitive RMSEs in the range of 0.62 to 0.68. Lower performance is observed from GAT~\cite{xiong2019pushing} and GSL-MPP~\cite{gslmpp}, with RMSE just over 0.70. In contrast, Weave~\cite{kearnes2016molecular} and AttentiveFP~\cite{attentivefp} yield weaker outcomes, particularly AttentiveFP with a low PCC of $0.2992 \pm 0.0816$. MPNN~\cite{gilmer2017neural} performs the worst, with the highest RMSE ($0.7850 \pm 0.0387$) and a negative PCC ($-0.0400 \pm 0.1133$), indicating poor generalization.

These results underscore the importance of incorporating geometry-informed atomic representations. The superior performance of GMC-MPNN validates the role of WCS in enhancing the model’s ability to generalize to novel molecules and accurately predict BBB permeability.

\section{Quantifying the Impact of Atom-Pair Interactions in BBBP Prediction}

To understand the contribution of different long-range geometric interactions within our GMC-MPNN model, we conducted an ablation study on the $BBBP_{\text{cls}}^{\text{MolNet}}$ dataset. In this study, we systematically removed the influence of specific atom-pair interactions from the model's graph construction process and re-evaluated its performance. By measuring the change in the AUC-ROC score, we can quantify the importance of each interaction for the blood-brain barrier permeability prediction task. The model's performance change was evaluated over 5 runs with different random seeds to ensure statistical robustness.

\begin{figure}[h!]
    \centering
    \includegraphics[width=0.9\textwidth]{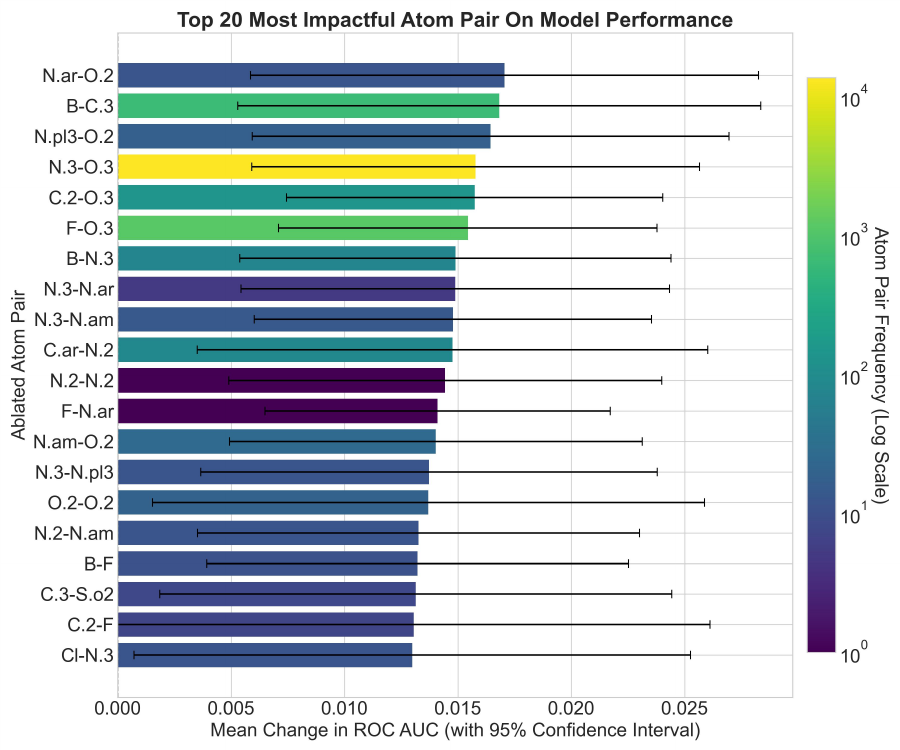} 
    \caption{Impact of ablating the top 20 atom-pair interactions on GMC-MPNN performance on the $BBBP_{\text{cls}}^{\text{MolNet}}$ dataset. Bars represent the mean change in AUC-ROC score over 5 seeds, with error bars showing the 95\% confidence interval. The color of each bar corresponds to the total frequency of that atom pair in the dataset, plotted on a logarithmic scale as indicated by the colorbar. A positive change indicates improved performance upon removal of the interaction.}
    \label{fig:ablation_study}
\end{figure}

The results of this study are summarized in Figure \ref{fig:ablation_study}, which illustrates two key properties for the top 20 atom-pair interactions whose removal most significantly impacted model performance. The length of each bar indicates the mean increase in the AUC-ROC score upon ablation, while the color of the bar represents the total frequency of that pair in the dataset, presented on a logarithmic scale.

Fig. \ref{fig:ablation_study} indicates that there appears to be no direct correlation between an interaction's frequency and the performance gain observed upon its removal. The top 20 includes both relatively rare interactions (indicated by darker, purple-blue colors) and more common ones (brighter, yellow-green colors), all of which yield a similar, statistically significant performance increase between approximately 0.011 and 0.018. The narrow 95\% confidence intervals for these changes are all firmly in positive territory, confirming the reliability of this effect.

Fig. \ref{fig:ablation_study} also reveals that many of the impactful interactions involve heteroatoms such as N, O, F, Cl, and B in hybridized or aromatic forms  (N.ar-O.2, N.pl3-O.2, F-O.3, Cl-N.3, etc.). These atom pairs reflect functional motifs relevant to hydrogen bonding, polarity, and halogen interactions, which are all known to influence blood-brain barrier permeability

Interestingly, some of the most frequent atom-pair interactions, such as N.3-O.3, exhibit substantial changes in AUC upon removal. This suggests that these interactions are not merely common, but core contributors to the model's understanding of BBBP. Their prevalence likely reflects the ubiquity of functionally important motifs, such as amine–hydroxyl or amine–ether interactions, which are associated with molecular solubility, polarity, and hydrogen bonding. They are the key determinants of passive diffusion across the BBB. The strong model dependence on these interactions underscores the importance of accurately encoding and learning from high-frequency, chemically relevant features, rather than assuming their impact is redundant or saturated.

Conversely, several low-frequency interactions, such as C.3-S.o2, Cl-N.3, and O.2-O.2, also result in large AUC changes when ablated. This indicates that rare but chemically specific substructures can influence a disproportionately large effect on prediction, either by encoding key pharmacophoric features or by introducing outlier geometries that heavily influence decision boundaries. This further supports the need for fine-grained subgraph modeling that balances both common and rare interactions based on chemical informativeness, not frequency alone.


\section{Conclusion}
In this work, we introduce GMC-MPNNs, a novel framework designed to advance the predictive modeling of BBBP. Unlike prior graph-based models that focus primarily on molecular connectivity, GMC-MPNNs explicitly incorporate spatial geometry and long-range atom-pair interactions through WCS. This approach enables a more nuanced encoding of atomic environments by integrating structural, geometric, and chemical context into node and edge representations.

Our comprehensive evaluation across three benchmark datasets, including BBBP$^\text{MolNet}_{\text{cls}}$, BBBP$^\text{B3DB}_{\text{cls}}$, and BBBP$^\text{B3DB}_{\text{reg}}$, shows that GMC-MPNN consistently outperforms state-of-the-art models in both classification and regression tasks. Notably, it achieves superior AUC-ROC scores on challenging, scaffold-balanced datasets. It also delivers lower RMSE and higher correlation coefficients in continuous logBB prediction. These results highlight the model's robust generalization ability, particularly when dealing with structurally diverse and previously unseen molecules.

Through an in-depth ablation analysis, we further demonstrate the significance of specific atom-pair interactions in influencing prediction accuracy. The results demonstrate the model's capacity to extract chemically meaningful features, including both frequent and rare atomic motifs. These motifs correlate with functional roles in BBB transport mechanisms, such as hydrogen bonding and polarity. Future directions may include expanding the framework to multi-modal biological datasets, incorporating protein-ligand interactions for transporter-based modeling, and adapting the architecture to real-world pharmacokinetic endpoints.

\section*{Acknowledgements}
This work is supported in part by funds from the National Science Foundation (NSF: \# 2516126, \# 2151802, and \# 2534947).
\section*{Availability of data and materials}
The source code is available at the GitHub repository: \url{https://github.com/MathIntelligence/GMC-MPNN-BBBP}.











\bibliographystyle{unsrt}
\bibliography{refs}

\end{document}